%% file: ecai-sample-and-instructions.tex
\documentclass{ecai}

\usepackage{latexsym}
\usepackage{amssymb}
\usepackage{amsmath}
\usepackage{amsthm}
\usepackage{booktabs}
\usepackage{enumitem}
\usepackage{graphicx}
\usepackage{color}
\usepackage{algpseudocode}
\usepackage{txfonts}
\usepackage{tabularx}    %
\usepackage{colortbl}    %
\usepackage{xcolor}      %
\usepackage{booktabs}    %
\usepackage{float}    %

\usepackage{tikz}
\usepackage{graphicx}
\usepackage{subcaption}
\usetikzlibrary{arrows.meta, positioning, fit, backgrounds, calc}

\newcommand{\BibTeX}{B\kern-.05em{\sc i\kern-.025em b}\kern-.08em\TeX}

\begin{document}

\begin{frontmatter}

\paperid{2978}

\title{Net-Zero: A Comparative Study on Neural Network Design for Climate-Economic PDEs Under Uncertainty}

\author[A,B,C]{\fnms{Carlos}~\snm{Rodriguez-Pardo}\orcid{0000-0001-6121-7738}\thanks{Corresponding Author. Email: carlos.rodriguezpardo.jimenez@gmail.com.}}
\author[A,B,C]{\fnms{Louis}~\snm{Daumas}\orcid{0000-0003-3239-8538}}
\author[A,B,C]{\fnms{Leonardo}~\snm{Chiani}\orcid{0009-0007-2491-6290}} 
\author[A,B,C]{\fnms{Massimo}~\snm{Tavoni}\orcid{0000-0001-5069-4707
}} 

\address[A]{Politecnico di Milano}
\address[B]{Euro-Mediterranean Center on Climate Change (CMCC)}
\address[C]{RFF-CMCC European Institute on Economics and the Environment (EIEE)}

\begin{abstract}
Climate-economic modeling under uncertainty presents significant computational challenges that may limit policymakers' ability to address climate change effectively. This paper explores neural network-based approaches for solving high-dimensional optimal control problems arising from models that incorporate ambiguity aversion in climate mitigation decisions. We develop a continuous-time endogenous-growth economic model that accounts for multiple mitigation pathways, including emission-free capital and carbon intensity reductions. Given the inherent complexity and high dimensionality of these models, traditional numerical methods become computationally intractable. We benchmark several neural network architectures against finite-difference generated solutions, evaluating their ability to capture the dynamic interactions between uncertainty, technology transitions, and optimal climate policy. Our findings demonstrate that appropriate neural architecture selection significantly impacts both solution accuracy and computational efficiency when modeling climate-economic systems under uncertainty. These methodological advances enable more sophisticated modeling of climate policy decisions, allowing for better representation of technology transitions and uncertainty—critical elements for developing effective mitigation strategies in the face of climate change.
\end{abstract}

\end{frontmatter}

\input{sections/introduction}

\input{sections/related_work}

\input{sections/overview}

\input{sections/methodology}

\input{sections/results}
\input{sections/conclusions}

\begin{ack}
This project received funding from the European Union European Research Council (ERC) Grant project No 101044703 (EUNICE)
\end{ack}

\newpage
\bibliography{mybibfile}

\end{document}

%% file: sections/introduction.tex
\section{Introduction}

Climate change represents one of humanity's most pressing challenges, demanding robust policy responses. Integrated assessment models (IAMs) that combine climate and economic systems have become essential tools for evaluating mitigation strategies. Yet, they face significant computational and conceptual challenges when incorporating realistic uncertainty related to climate impacts and mitigation processes. These uncertainties are particularly pronounced when considering disruptive decarbonization technologies whose future availability and effectiveness are unknown.

IAMs often rely on deterministic approaches or simple uncertainty representations, failing to capture the complex interplay between decision-making under uncertainty and technological transitions. Recently, sophisticated stochastic models incorporating ambiguity aversion and model misspecification have emerged. These better represent the challenges faced by policymakers who must act without knowing probability distributions governing climate and technological dynamics. However, the high-dimensional partial differential equations (PDEs) that characterize these models quickly become computationally intractable using numerical methods.

Recent advances in deep learning provide promising directions for addressing these challenges. Neural network-based PDE solvers have demonstrated remarkable capabilities in handling high-dimensional problems that would be infeasible for traditional methods. However, their application to climate economics models with ambiguity, and discrete technological jumps remains unexplored. Different neural architectures may vary in their ability to capture the unique characteristics of these problems, such as discontinuities in policy functions around technological breakthroughs or the complex effects of uncertainty aversion on optimal control. Therefore, we argue that there is a lack of understanding of the design choices that govern the performance of neural PDE solvers in climate and economic applications. This gap may lead to inaccurate solutions or hinder the possibility of achieving high-performing, scalable models in this field. 

Our main goal in this paper is, therefore, to explore the design space of neural PDE solvers for climate-economic modeling under uncertainty. These models face unique computational challenges including high dimensionality and non-linear interactions between economic and climate variables. Leveraging a \textit{ground-truth} dataset generated using a Finite-Differences solver, we explore the impact of several important neural network design factors, such as architecture types, residual connections, activation functions, optimization algorithms, and regularization techniques. We find that commonly used approaches often provide suboptimal trade-offs between computational efficiency and accuracy, hindering progress in the field and limiting the application of neural PDE solvers for climate-economic policy design. By applying targeted modifications to these models, we demonstrate significant improvements in both accuracy and computational efficiency. We present the following contributions:

\begin{itemize}
    \item A continuous-time endogenous-growth economic model incorporating mitigation options under uncertainty.
    \item A systematic evaluation of neural architectures for solving the resulting PDEs, comparing their accuracy and computational efficiency against finite-difference  solutions. 
    \item New insights into how neural network design affect the ability to model optimal climate policy under uncertainty.
    \item An open source implementation and data made available upon publication,  enabling further research at the intersection of climate economics and machine learning.
\end{itemize}
\vspace{-3mm}

%% file: sections/related_work.tex
\begin{figure*}[t]
\centering
\begin{tikzpicture}[
    node distance=0.5cm and 0.8cm,
    box/.style={
        draw=black!70,
        fill=blue!10,
        thick,
        minimum width=3.5cm,
        minimum height=2cm,
        rounded corners=2pt,
        align=center
    },
    smallbox/.style={
        box,
        minimum height=1.5cm
    },
    arrow/.style={
        thick,
        -Stealth,
        black!70
    },
    highlight/.style={
        rounded corners=3pt,
        draw=blue!40,
        dashed,
        thick,
        inner sep=4pt
    }
]

\node[box] (econ) {{\bf Economic Model}\\\small • Multiple mitigation options\\\small • Ambiguity aversion};

\node[box, right=of econ] (hjb) {{\bf HJB Equation (Sec.~\ref{sec:econ_model})}\\\small • Value function\\\small • Optimal control\\\small • Uncertainty adjustments};

\node[box, right=of hjb, minimum width=6.5cm] (neural) {{\bf Neural Architectures (Sec.~\ref{sec:nn_design})}\\
    \begin{tabular}{cc}
    \small DGM Variants & \small SIREN \\
    \small Residual Networks & \small FNO \\
    \small Recurrent Networks & \small Self-Attention
    \end{tabular}
};

\node[smallbox, below=of hjb] (gt) {{\bf \textit{Ground Truth} (Sec.~\ref{sec:linear_solver})}\\\small Finite-Difference Solution};

\node[smallbox, below=of econ] (eval) {{\bf Evaluation (Sec.~\ref{sec:gt_val})}\\\small • Value and policy errors\\\small • Computational efficiency};

\node[smallbox, below=of neural, minimum width=6.5cm] (result) {{\bf Optimal Architecture (Sec.~\ref{sec:results})}\\
    \begin{tabular}{cc}
    \small Value Network & \small Policy Network
    \end{tabular}
};

\draw[arrow] (econ) -- (hjb);
\draw[arrow] (hjb) -- (neural);
\draw[arrow] (hjb) -- (gt);
\draw[arrow] (gt) -- (eval);
\draw[arrow] (neural) -- (result);
\draw[arrow] (eval.south) -- ++(0,-0.5) -| (result.south);

\node[highlight, fit={($(neural.north west)+(-0.1,0.1)$) ($(result.south east)+(0.1,-0.1)$)}] (highlight) {};

\end{tikzpicture}
\caption{Our neural PDE method for climate-economic models under uncertainty. Our approach builds on a climate-economic model, formulated as a Hamilton-Jacobi-Bellman equation. We compare various neural architectures within a two-network framework against finite-difference ground truth solutions.}
\label{fig:method}
\end{figure*}
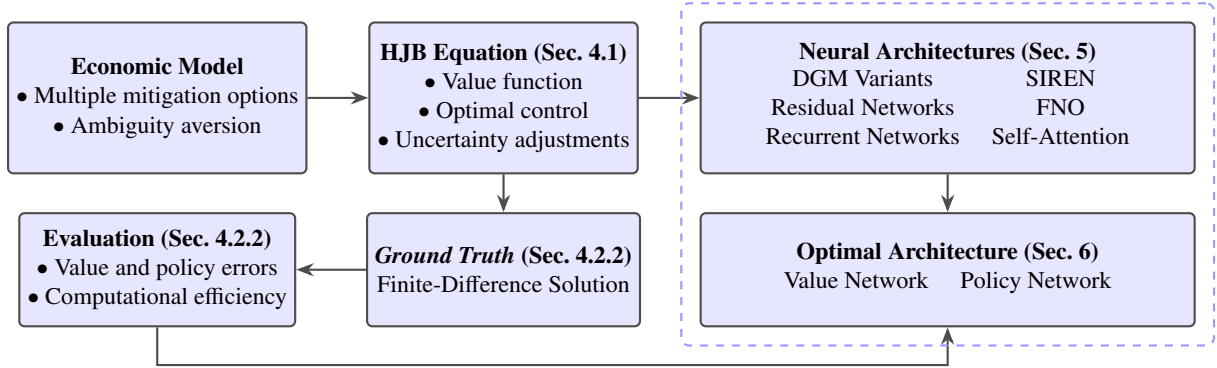

\section{Related Work}
Our work intersects climate economics under uncertainty, neural approaches to solving partial differential equations (PDEs), and computational methods for climate policy analysis. We build upon key advances in these domains while addressing important limitations.

\subsection*{Climate Economics under Uncertainty}

Climate-economic models increasingly incorporate uncertainty to better represent real-world decision contexts. While early contributions used deterministic integrated assessment models (IAMs) like DICE \cite{nordhaus1992optimal}, more recent approaches have employed stochastic extensions \cite{cai2018dsice, lontzek2015stochastic} to account for climate risk. Beyond this, ambiguity aversion frameworks have been developed where decision-makers face unknown probability distributions \cite{millner2013scientific, lemoine2016ambiguous}.  The Hansen-Sargent model misspecification framework \cite{hansen2008robustness} offers a rigorous approach to handling deep uncertainty by optimizing against worst-case scenarios. Recent applications by \citet{barnett2020pricing} and \citet{barnett2021climate} to climate economics demonstrate its impact on the social cost of carbon and financial markets. \citet{berardi2023uncertainty} further explored how uncertainty and sentiment affect asset prices in this framework. Our work extends these models by incorporating mitigation technologies and explicitly modeling uncertain technological jumps—responding to calls from \citet{grant2021confronting} about the need to better represent mitigation deterrence under uncertainty.

\subsection*{Neural Networks as PDE Solvers}

Traditional numerical methods for PDEs face significant challenges in high-dimensional settings due to the curse of dimensionality. The Physics-Informed Neural Networks framework \cite{raissi2019physics} introduced neural approaches for solving PDEs, while the Deep Galerkin Method (DGM) \cite{sirignano2018dgm} pioneered a mesh-free approach using domain sampling. \citet{al-aradi2022} extended this approach with their Policy Iteration Algorithm (DGM-PIA), which simultaneously optimizes value functions and control policies in an adversarial framework. Recent advances have explored specialized architectures for PDE solving, including periodic activations in SIREN \cite{sitzmann2020implicit}, spectral approaches in Fourier Neural Operators \cite{li2021fourier}, attention mechanisms \cite{cao2021choose}, and various residual connections strategies \cite{he2016deep, srivastava2015highway}. However, as noted by \citet{cuomo2022scientific}, comprehensive comparisons of neural architectures for climate-economic PDEs remain scarce. \citet{huang2023partial} provides a recommendable survey on the topic, but lacks application-specific guidance for complex economic systems.

\subsection*{Computational Challenges in Climate Policy}

The computational complexity of climate-economic models under uncertainty has limited their application in policy settings. Leading models either use simplified representations with few states \cite{dietz2021economic, fillon2023optimal} or employ approximations \cite{heutel2022climate} that may miss critical interactions.  Neural PDE solvers offer promising alternatives, but their application to climate economics has focused primarily on baseline architectures. \citet{barnett2023deep} demonstrated the power of neural methods for climate policy analysis but did not explore architectural variations. \citet{anagnostou2023machine} surveyed machine learning for climate policy models, highlighting the need for more rigorous benchmarking. Our methodological contribution fills this gap by systematically evaluating neural architectures for climate-economic PDEs with mitigation options. Deep network architectures have also been used to tackle broader types of uncertainty. \citet{friedl2023deep} developed the Deep Equilibrium Network methodology to run sensitivity analysis on key components of stochastic IAMs.
Our work advances the state-of-the-art by: (1) developing a richer climate-economic model incorporating multiple mitigation pathways and technological uncertainty; (2) systematically evaluating neural architectures against finite-difference benchmarks; and (3) providing insights into how architectural choices affect solution quality for climate policy applications. This approach enables more sophisticated modeling of technology transitions under uncertainty—addressing a critical need in climate policy design \cite{kriegler2018pathways, grant2021confronting}.

%% file: sections/overview.tex
\section{Overview}
Our paper examines neural architecture design for climate-economic PDEs under uncertainty. We illustrate our method in Figure~\ref{fig:method}. We structure the rest of the paper as follows. 
In Section~\ref{sec:econ_model}, we present our economic model with multiple mitigation options and explicit uncertainty representation, deriving the resulting Hamilton-Jacobi-Bellman equation. Next, in Section~\ref{sec:fd_approach}, we formulate the specific PDEs arising from this economic framework, highlighting the mathematical properties that challenge numerical solvers and motivate our architectural exploration. We describe our finite-difference implementation in Section~\ref{sec:gt_val}, which we use to generate \textit{ground-truth} solutions for benchmarking.
The methodological contributions appear in Section~\ref{sec:nn_design}, where we systematically present neural network design that address the specific characteristics of our climate-economic PDEs. Section~\ref{sec:hyperopt} briefly outlines our optimization strategy and computational implementation, followed by Section~\ref{sec:results} which presents our comparative analysis of architectural performance, and shows how these methodological advances translate to enhanced capabilities for analyzing climate policy questions under uncertainty, with Section~\ref{sec:conclusion} summarizing key insights and future directions.

%% file: sections/methodology.tex
\section{Methods}
In this section, we formalize the climate-economic model we will use for our experiments (Sec.~\ref{sec:econ_model}), our finite-differences solution (Sec.~\ref{sec:fd_approach}), and how we compute the benchmark metrics (Sec~\ref{sec:gt_val}).

\subsection{Economic model}\label{sec:econ_model}

For our climate-economic model, we consider a modified version of the continuous-time stochastic economic framework proposed by \cite{barnett_deep_2023}. We choose this model for its relative simplicity, which makes it feasible to compute solutions using traditional Finite Differences methods, key for robust benchmarking of neural solvers. 
This model is composed of four states variables: $k$, the logarithm of the capital stock ($K$), $S_L$, the share of low-carbon capital in the economy, $\Gamma$ the temperature anomaly, and $n$ the logarithm of climate damage intensity ($N$). We assume two controls, $i_L$ and $i_H$, the investment rates in low and high-carbon capital respectively. This problem is solved assuming that a benevolent social planner with rational expectations (\textit{i.e.} optimizes in expectations while correcting for stochasticity when needed) maximizes welfare given by:
\begin{equation}
\label{eq:maxprob}
V = \int_0^{\infty}e^{-\rho}u(C)dt = \int_0^{\infty}e^{-\rho}\log(C)dt.
\end{equation}
In the equation, $\rho$ is the discount rate and $C$ is consumption given by:
\begin{equation}
C = Y - I,
\end{equation}
with $Y$ being the production and $I$ the investment. $Y$ is given by $Y =   \alpha K$. A fraction $S_L$ of the capital $K$ is made of low-carbon capital, such that emissions are given by:
\begin{equation}
E = \lambda\alpha K(1 - S_L),
\end{equation}
where the parameter $\lambda$ is the carbon intensity. Emissions feed into temperature anomaly $\Gamma$ with a sensitivity $\zeta$ and a stochastic term:
\begin{equation}
d\Gamma = E(\zeta  + \sigma_\Gamma dW).
\end{equation}
The economy suffers from damage $N$, whose logarithm $n$ evolves with the temperature anomaly:
\begin{equation}
\tfrac{dn}{d\Gamma} = \eta_0 + \eta_1\Gamma.
\end{equation}
By applying Itô's Lemma:
\begin{equation}
dn = \left(\eta_0 + \eta_1 d\Gamma\right)d\Gamma + \tfrac{1}{2}\eta_1\sigma_\Gamma^2E^2.
\end{equation}
To avoid damage, the social planner chooses between high and low-carbon investment, with $i_L$, $i_H$ are the investment rates.:
\begin{equation}
I = I_L + I_H = \left[i_LS_L + i_H(1-SL)\right]K.
\end{equation}
We assume quadratic adjustment costs to investment. Hence, the law of motion of $k$ is given by:\footnote{This law of motion can be easily retrieved from the following law of motion in levels for aggregate capital: $dK = (-\delta + I - \kappa I^2)$ by considering $I = I_L + I_H = (i_LS_L + i_H(1-SL))$ and $K = K_L + K_H =S_L K + (1- SL)K$, Where $K_L$ is low-carbon capital and $K_H$ high carbon capital. }
\begin{align}
dk &= S_L\left(-\delta_L + i_L - \kappa_Li_L^2 + \tfrac{1}{2}(\sigma_LSL)^2\right) + \\
&(1-S_L)\left(-\delta_H + i_H - \kappa_Hi_H^2 + \tfrac{1}{2}(\sigma_H(1-SL))^2\right) + \nonumber \\
& S_L\sigma_LW_L + (1-S_L)\sigma_HW_H \nonumber \\
& = S_L dk_L + (1-S_L)dk_H. \nonumber
\end{align}
Where, for $j\in{L,H}$, $\kappa_j$ is the coefficient for adjustment costs, $\sigma_j$ a volatility parameter, $W_j$ Wiener processes, and $dk_j$ is the law of motion of the log-value of capital $j$. Note that as per Îtô's Lemma, the laws of motion in level would write:
\begin{align}
dK_L &= (-\delta_L + i_L - \kappa_Li_L^2) + \tfrac{1}{2}(\sigma_LSL) \sigma_LW_L,  \\
dK_H &= (-\delta_H + i_H - \kappa_Hi_H^2) + \tfrac{1}{2}(\sigma_HSH) \sigma_HW_H.  
\end{align}
The law of motion for $S_L$ writes as follows:\footnote{Again, considering $K = K_L + K_H = \tfrac{K_L}{K}K + \tfrac{K_H}{K} = S_L K + (1- SL)K$, one can write $dS_L = S_L(dK_L - dK)$.  As per the definitions in the body of the text, it is easy do show that $dS_L = S_L(dK_L - S_LdK - (1-S_L)dK = S_L(1-S_L)(dK_L-dK_H)$, that can be turned into logs applying Itô's Lemma.}
\begin{align}
dS_L &= S_L(1-S_L)(-\delta_L + i_L - \kappa_Li_L^2 + \tfrac{1}{2}(\sigma_LSL)^2 - \\
&(-\delta_H + i_H - \kappa_Hi_H^2 + \tfrac{1}{2}(\sigma_H(1-SL))^2 ) + \nonumber \\
&\tfrac{1}{2}S_L(1-S_L)(\sigma_LW_L + \sigma_HW_H). \nonumber
\end{align}
We assume that damage affect the capital stock, investment and consumption alike, such that all these quantities are divided by $N$:
\begin{equation}
\Tilde{C} = \tfrac{C}{N},
\end{equation}
such that:
\begin{equation}
U(\Tilde{C}) = \log(\tfrac{C}{N}) = u(C) - n,
\end{equation}
\begin{equation}
\log(\Tilde{K}) = \log\left(\tfrac{K}{N}\right) = \log(K) - \log(N) = k - n.
\end{equation}

\noindent Based on the above, the maximization problem in Equation \ref{eq:maxprob} can be reformulated recursively as a Hamilton-Jacobi-Bellman (HJB) equation. We define the following shorthands:
\begin{align}
dk^* &= dk - S_L\sigma_LW_L + (1-\S_L)\sigma_HW_H,  \\
dS_L^* &= dS_L - \tfrac{1}{2}S_L(1-S_L)(\sigma_LW_L + \sigma_HW_H), \\ 
d\Gamma^* &= d\Gamma - \sigma_\Gamma dW, \\
dn^*      &= dn - (\eta_0 + \eta_1 d\Gamma)\sigma_\Gamma dW,
\end{align}
and write the HJB problem as:
\begin{align}
    \rho V &= \rho\log\left(\tfrac{\alpha K - i_LS_L - i_H(1-S_L)}{N}\right) + \\
            & \tfrac{1}{2}\tfrac{\partial V}{\partial k^2}\left((\sigma_LS_L)^2 + (\sigma_H(1-S_L))^2\right) +\nonumber \\
            & \tfrac{\partial V}{\partial k}dk^* +  \tfrac{\partial V}{\partial S_L}dS_L^*+ \nonumber \\
            & \tfrac{1}{2}\tfrac{\partial V}{\partial S_L^2}\left((S_L(1-S_L))^2(\sigma_L^2 + \sigma_H^2)\right)+ \nonumber\\
            & \tfrac{\partial V}{\partial \Gamma}d\Gamma^* +  \tfrac{1}{2}\tfrac{\partial V}{\partial \Gamma^2}E^2\sigma_\Gamma^2  +\nonumber\\
            &\tfrac{\partial V}{\partial n}dn^* + \tfrac{1}{2}\tfrac{\partial V}{\partial n^2}(\eta_0 + \eta_1\Gamma)^2E^2\sigma_\Gamma^2 .\nonumber\\ \nonumber
\end{align}

\noindent This expression can be further simplified by verifying that $V = v - n$ (see \cite{barnett_climate_2021}). As a result, the problem is reduced to a three-state HJB, in which $\frac{\partial V}{\partial n} = -1$ and $\frac{\partial V}{\partial n^2} = 0$:
\begin{align}
    \rho v&= \rho\log\left(\tfrac{\alpha K - i_LS_L - i_H(1-S_L)}{N}\right) \\
            & \tfrac{\partial V}{\partial k}dk^* +\tfrac{\partial V}{\partial S_L}dS_L^* +\nonumber \\
            & \tfrac{1}{2}\tfrac{\partial V}{\partial k^2}\left((\sigma_LS_L)^2 + (\sigma_H(1-S_L))^2\right) +\nonumber \\
            & \tfrac{1}{2}\tfrac{\partial V}{\partial S_L^2}\left((S_L(1-S_L))^2(\sigma_L^2 + \sigma_H^2)\right) +\nonumber\\
            & \tfrac{\partial V}{\partial \Gamma}d\Gamma^*+  \tfrac{1}{2}\tfrac{\partial V}{\partial \Gamma^2}E^2\sigma_\Gamma^2  -dn^*. \nonumber 
\end{align}
This gives us a partial-differential equation that we will aim to approximate with various neural network architectures.

\subsection{Finite-difference approach to PDEs}\label{sec:fd_approach}

A standard approach to PDE solving is the use of finite-difference methods. We solve this problem with this more traditional approach to provide ground truth to evaluate our neural networks. We use the same method as \cite{barnett_climate_2021, barnett_deep_2023}.%
Considering the three-state problem defined above, we define a linear grid along the three states $k$, $S_L$ and $\Gamma$, with points separated by spaces $h_k$, $h_{SL}$ and $h_\Gamma$. For each point of the grid, we compute the gradient of the value function, derive controls, and update the value function accordingly by solving for it with a linear solver. We iterate these operations until convergence. Our algorithm is summarized as follows, and we detail the steps below.
\begin{algorithmic}
\State \textbf{Input}: Initial guess for value function $v_0$, $\epsilon=1e^{-8}$ \\
Initialize n = 0, $v_0^n = v_0$
\While{$|v_0^{n+1} - v_0^n| > \epsilon$} \\
Step 1: Solve for optimal controls $\{i_L^{n+1}, i_H^{n+1}\}$ \\
Shallow approach applied \\
Step 2: Solve for value function \\
Linear solver (implicit scheme) applied \\
Step 3: Check for convergence 
\If{$|v_0^{n+1} - v_0^n| < \epsilon$} stop, otherwise continue  
\EndIf
\EndWhile
\State \textbf{Return} $v_0^n$
\end{algorithmic}

\subsubsection{Policy iterations}

\paragraph{Gradient computation}
To derive the controls, we first compute the gradient of our value function along the grid. For interior points, we use a central-difference scheme with natural boundary conditions on the edges of the domain. For a state $x$, with $i$ the indexation of state values along the grids, $h_x$ the corresponding grid step and $\ell$ the iteration index:
\begin{equation}
\left(\tfrac{\partial V^{(\ell)}}{\partial x}\right)_i^{(\ell)} = \tfrac{f_{(i+1)} - f_{(i-1)}}{2h_x},
\end{equation}
\begin{equation}
\left(\tfrac{\partial^2 V^{(\ell)}}{\partial x^2}\right)_i^{(\ell)} = \tfrac{f_{(i+1)} + f_{(i-1)} - 2f_{(i)}}{h_x^2}.
\end{equation}

\noindent Central difference schemes are widely used given their stabilising property. For a lower-boundary point, we use a forward difference:
\begin{equation}
\left(\tfrac{\partial V^{(\ell)}}{\partial x}\right)_0^{(\ell)} = \tfrac{f_{(1)} - f_{(0)}}{h_x},
\end{equation}
\begin{equation}
\left(\tfrac{\partial V^{(\ell)}}{\partial x^2}\right)_0^{(\ell)} = \tfrac{f_{(2)} + f_{(0)} - 2f_{(1)}}{h_x^2}.
\end{equation}
For a higher-boundary point, we use a backward difference:
\begin{equation}
\left(\tfrac{\partial V^{(\ell)}}{\partial x}\right)_{I-1}^{(\ell)} = \tfrac{f_{(I-1)} - f_{(I-2)}}{h_x},
\end{equation}
\begin{equation}
\left(\tfrac{\partial V^{(\ell)}}{\partial x^2}\right)_{I-1}^{(\ell)} = \tfrac{f_{(I-1)} + f_{(I-3)} - 2f_{(I-2)}}{h_x^2}.
\end{equation}

\paragraph{First-order conditions}
We first consider the set of previous controls and derive the marginal value of consumption:
\begin{equation}
mc = \rho\tfrac{\partial\log(C)}{\partial C} = \nonumber\tfrac{\rho}{(\alpha K - S_Li_L^{(\ell)}K - (1-S_L)i_H^{(\ell)}K}.
\end{equation}

\noindent We then derive the first-order conditions for the two controls, that determine their optimal values given the set of gradients:
\begin{equation}
\left(i_L^{(\ell+1)}\right)^* = \tfrac{1}{2\kappa_L}\left(1- \tfrac{mc}{\frac{\partial V^{(\ell)}}{\partial k} - (1-S_L)\frac{\partial V^{(\ell)}}{\partial S_L}}\right),
\end{equation}
\begin{equation}
\left(i_H^{(\ell+1)}\right)^* = \tfrac{1}{2\kappa_H}\left(1- \tfrac{mc}{\tfrac{\partial V^{(\ell)}}{\partial k} - S_L\tfrac{\partial V^{(\ell)}}{\partial S_L}}\right).
\end{equation}

\noindent For stability, we update the controls with a relaxation parameter $\chi$:
\begin{equation}
i_L^{(\ell+1)} = \chi\left(i_L^{(\ell+1)}\right)^* + (1-\chi)i_L^{(\ell)},
\end{equation}
\begin{equation}
i_H^{(\ell+1)} = \chi\left(i_H^{(\ell+1)}\right)^* + (1-\chi)i_H^{(\ell)}.
\end{equation}

\noindent We update the value of consumption and of the laws of motion and proceed to the linear solver. We also implemented a Cobweb algorithm which, instead of directly updating the controls, starts with a first guess on the control and updates the marginal value of consumption, iterating until convergence. No differences were found across solutions, we thus proceed with the shallower approach.

\subsubsection{Linear solver}\label{sec:linear_solver}

We compact the HJB equation as:
\begin{equation}
-A\rho v + \tfrac{\partial V}{\partial k}B_k + \tfrac{\partial V}{\partial S_L}B_{SL}  + \tfrac{\partial V}{\partial \Gamma}B_{\Gamma}  + \tfrac{\partial V}{\partial k^2}C_k + \tfrac{\partial V}{\partial S_L}C_{SL}  + \tfrac{\partial V}{\partial \Gamma}C_{\Gamma} + D = 0,
\end{equation}

\noindent With $ A = \mathbf{I}_I$, and where $\mathbf{I}_I$ is the identity matrix. The first-order partial derivative coefficients are:
\begin{align}
B_k = \text{ }& S_L(-\delta_L + i_L^{(\ell+1)} - \kappa_L(i_L^{(\ell+1)})^2 + \\ 
& \tfrac{1}{2}(\sigma_L S_L)^2 + \nonumber\\
& (1-S_L)(-\delta_H + i_H^{(\ell+1)}  - \kappa_H(i_H^{(\ell+1)})^2 + \nonumber \\
& \tfrac{1}{2}(\sigma_H(1-S_L))^2, \nonumber
\end{align}
\begin{align}
B_{SL} =  \text{ }&S_L(1-S_L)(-\delta_L + i_L^{(\ell+1)} - \kappa_L(i_L^{(\ell+1)})^2 + \\
& \tfrac{1}{2}(\sigma_LSL)^2 - \nonumber\\
& (-\delta_H + i_H^{(\ell+1)}  - \kappa_H(i_H^{(\ell+1)})^2 + \nonumber \\
& \tfrac{1}{2}(\sigma_H(1-SL))^2 )).  \nonumber
\end{align}
\begin{align}
B_{\Gamma} &=  E\zeta
\end{align}

\noindent The second order partial derivative coefficients are:
\begin{align}
C_{k} &=   \tfrac{1}{2}\left((\sigma_LS_L)^2 + (\sigma_H(1-S_L))^2\right), \\
C_{SL} &=   \tfrac{1}{2}\left((S_L(1-S_L))^2(\sigma_L^2 + \sigma_H^2)\right),\\
C_{\Gamma} &= \tfrac{1}{2}E^2\sigma_\Gamma^2.
\end{align}
The constant term is given by:
\begin{align}
 D &= \rho\log\biggl(\alpha K  - S_Li_L K - (1-S_L)i_H K\biggl) \\
 & - \biggl( (\eta_0 + \eta_1 d\Gamma)d\Gamma + \tfrac{1}{2}\eta_1\sigma_\Gamma^2E^2\biggl).    \nonumber
\end{align}

\begin{table*}[htbp!]
\footnotesize
\renewcommand{\arraystretch}{1.3}
\begin{tabularx}{\textwidth}{|l|X|X|}
\hline
\rowcolor[HTML]{E6F2FF} 
\textbf{Architecture} & \textbf{Key Advantages} & \textbf{Potential Strengths for Climate-Economic PDEs} \\
\hline
DGM Variants & 
Foundational approach with proven success in PDE solving; mesh-free sampling enables scaling to higher dimensions & 
Effective baseline for HJB equations; explicitly designed to handle PDE residuals with non-uniform boundary conditions \\
\hline
\rowcolor[HTML]{F8F8F8} 
Residual Networks & 
Improved gradient flow in deep networks; allows for very deep architectures without vanishing gradients & 
Stable training across high-dimensional spaces; better preservation of low-level features important for policy decisions \\
\hline
Highway Networks & 
Adaptive information flow control; dynamically decides which information passes through the network & 
Better handling of varying sensitivities across state variables; can selectively focus on critical state regions \\
\hline
\rowcolor[HTML]{F8F8F8} 
Recurrent (LSTM/GRU) & 
Captures dependencies between variables; maintains internal memory through recurrent connections & 
May better model state-dependent transitions \\
\hline
SIREN & 
High-frequency component modeling; periodic activations capture oscillatory behavior & 
Represents sharp transitions around technological jumps; better approximation of discontinuities in policy functions \\
\hline
\rowcolor[HTML]{F8F8F8} 
FNO & 
Global pattern recognition; operates efficiently in frequency domain; learns spatial relationships & 
Model interactions across the full state space; captures multi-scale dynamics between climate and economy \\
\hline
Self-Attention & 
Dynamic relationship modeling; context-dependent feature weighting & 
Capturing climate-economic variable interactions; adaptively focuses on relevant state information \\
\hline
\end{tabularx}
\caption{Comparison of Neural Network Architectures for Climate-Economic PDE Solving.}
\label{tab:architecture_comparison}
\end{table*}

\noindent The linear solver is built in Eigen \citep*{eigenweb}, and uses an implicit scheme for better performances. For better accuracy, we derive the finite-difference solution over a $60^3$ grid, and with a tolerance of $10^{-8}$. Around 1500 iterations are needed to reach a solution, which takes approximately 2 hours to converge.

\subsubsection*{Validation against Ground Truth}\label{sec:gt_val}

This process provides us with \textit{Ground Truth}, which will consist of three outcomes: the value function $V^{**}$ and the controls $i_L^{**}$ and $i_H^{**}$. More precisely, our adversarial training framework will involve two deep networks. One $\mathcal{V}$ approximates the value function, while the other, $\mathcal{P}$, returns the controls, that we denote $\iota_L$ and $\iota_H$. To evaluate our architectures, we evaluate our trained models on the $60^3$ grid. We measure the geometric mean of absolute errors across the grid $\mathcal{G}$:
\begin{equation}
\mathcal{L}_{final} = \sqrt[3]{\mathcal{L}_{i_V} \times \mathcal{L}_{i_L} \times \mathcal{L}_{i_H}}  ,
\end{equation}
\begin{equation}
\mathcal{L}_{i_V} = \tfrac{1}{\mathrm{card}(\mathcal{G})}\sum_{i=1}^{\mathrm{card}(\mathcal{G})}|V_i -  \mathcal{V}(k, S_L, \Gamma)_i|,
\end{equation}
\begin{equation}
\mathcal{L}_{i_L} = \tfrac{1}{\mathrm{card}(\mathcal{G})}\sum_{i=1}^{\mathrm{card}(\mathcal{G})}|i_{L,i} -  \iota_L(k, S_L, \Gamma)_i|,
\end{equation}
\begin{equation}
\mathcal{L}_{i_H} = \tfrac{1}{\mathrm{card}(\mathcal{G})}\sum_{i=1}^{\mathrm{card}(\mathcal{G})}|i_{H,i} -  \mathcal{\iota}_H(k, S_L, \Gamma)_i|.
\end{equation}

\section{Neural Network Design}\label{sec:nn_design}

\begin{figure*}[ht!]
    \centering
    \includegraphics[width=\textwidth]{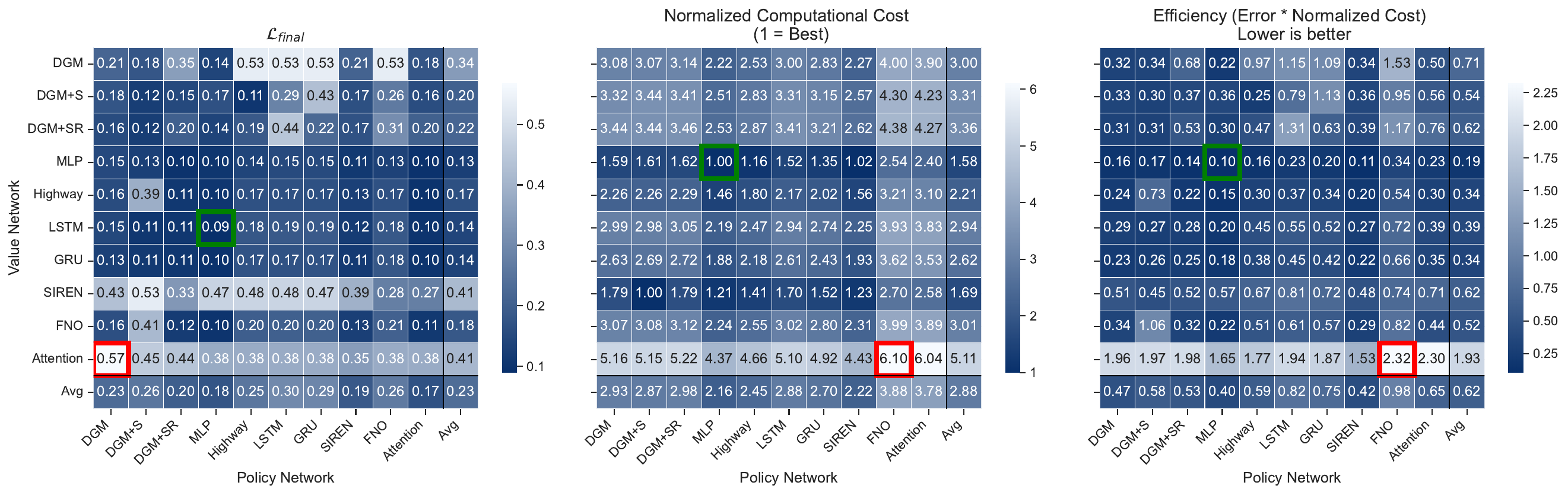}
    \caption{Performance comparison of different architecture combinations. On the left, we show the error with respect to our FD benchmark, in the middle, computational cost (1 is best, X means X times slower training times than the best case), and right is overall efficiency, combining error and computational cost. We highlight the best result in green, the worst in red. We also provide average results across rows and columns (\emph{avg}).}
    \label{fig:performance_comparison}
\end{figure*}

In this section, we describe our neural architectures for approximating solutions to the climate-economic PDE system presented before. Building upon recent advances in physics-informed neural networks for solving high-dimensional PDEs, we explore several architectural variants that address the specific challenges of our problem domain.

\subsection*{Two-Network Framework}

As described before, our  methodology employs two separate networks that work in tandem following an adversarial framework:
\begin{enumerate}
    \item A \textbf{value network} $\mathcal{V}_\theta(k, S_L, \Gamma)$ approximating the value function
    \item A \textbf{policy network} $\mathcal{P}_\phi(k, S_L, \Gamma)$ returning control variables $(i_L, i_H)$
\end{enumerate}
These networks are trained in advesarially, where the value network's loss is defined by the residual of the HJB equation, while the policy network is trained to maximize the expected value function subject to PDE dynamics. This approach is built upon the Deep Galerkin Method with Policy Iteration Algorithm (DGM-PIA) proposed by Al-Aradi et al.~\cite{al-aradi2022}, which has shown success in economic modeling. We explore a set of designs for the models ($\mathcal{V}_\theta$ AND $\mathcal{P}_\phi$ may be different architectures), which we also summarize in Table~\ref{tab:architecture_comparison}.

\subsection*{Base Architecture: Deep Galerkin Method}

Our foundation is the Deep Galerkin Method (DGM) introduced by Sirignano \& Spiliopoulos~\cite{sirignano2018}. The DGM approach employs deep neural networks trained to satisfy PDEs through stochastic sampling of the domain. A key advantage of this approach is its mesh-free nature, which helps overcome the curse of dimensionality inherent in our climate-economic model. We systematically explore several variants of the base DGM-PIA architecture:
\begin{enumerate}
    \item \textbf{DGM Baseline}: The original DGM architecture with TanH activations and non-residual complex hidden layers, matching the original implementation as closely as possible.

    \item \textbf{DGM with SiLU}: Replaces TanH with Sigmoid Linear Units activation functions~\cite{ramachandran2017searching}, defined as:
    \begin{equation}
        \text{SiLU}(x) = x \cdot \sigma(x),
    \end{equation}
    where $\sigma(x)$ is the logistic sigmoid function. These activations provide smoother gradients, which is particularly beneficial around the discontinuities caused by technological jumps in our model.

    \item \textbf{DGM with SiLU and Residual Connections}: Enhances the SiLU-based DGM with residual connections~\cite{he2016deep} between layers:
    \begin{equation}
        h_{i+1} = h_i + f_i(h_i),
    \end{equation}
    where $h_i$ is the output of layer $i$ and $f_i$ is a nonlinear transformation. Residual connections facilitate gradient flow during training, which enhances stability and fast convergence.
\end{enumerate}

\subsection*{Alternative Architecture Paradigms}

Beyond enhancing DGM variants, we explore fundamentally different architectural paradigms that may capture distinct mathematical properties of our climate-economic PDEs. For these models, we also employ SiLU activations where appropriate, and residual connections in every layer $f_i$. We implement and test the following models:

\begin{enumerate}
    \item \textbf{Residual and Highway Networks}:
    \begin{itemize}
        \item \textbf{Residual Networks}~\cite{he2016deep}: Uses standard residual connections with normalized activations, where each layer $f_i$ works as a shallow multi-layer perceptron. This architecture helps maintain gradient flow in very deep networks.
        
        \item \textbf{Highway Networks}~\cite{srivastava2015}: Employs gating mechanisms that modulate information flow:
        \begin{equation}
            h_{i+1} = t_i \odot h_i + (1 - t_i) \odot f_i(h_i),
        \end{equation}
        where $t_i$ is a learned  gate and $\odot$ denotes element-wise multiplication. These networks can adaptively control information flow, potentially helping with the non-linearities across our space.
    \end{itemize}

    \item \textbf{Recurrent Architectures}: These architectures can potentially better capture dependencies between state variables:
    \begin{itemize}
        \item \textbf{LSTM (Long Short-Term Memory)}~\cite{hochreiter1997}: Uses memory cells and gating structures to manage information flow.
        \item \textbf{GRU (Gated Recurrent Unit)}~\cite{cho2014}: Employs gating mechanisms with fewer parameters than LSTM, potentially improving efficiency.
    \end{itemize}

    \item \textbf{SIREN (Sinusoidal Representation Networks)}~\cite{sitzmann2020}: Employs periodic activations:
    \begin{equation}
        \sin(\omega_0 \cdot \mathbf{W}\mathbf{x} + \mathbf{b}).
    \end{equation}
    This architecture may better represent solutions with high-frequency components, which can emerge in our economic-climate system near transition boundaries.

    \item \textbf{FNO (Fourier Neural Operator)}~\cite{li2021fourier}: Implements a spectral approach that leverages the Fourier domain:
    \begin{equation}
        \mathbf{h}' = \text{FFT}^{-1}(R(\theta) \cdot \text{FFT}(\mathbf{h})),
    \end{equation}
    where $R(\theta)$ represents learnable filters in the Fourier domain. This approach is particularly suited for capturing both global and high-frequency patterns in the PDE solution.

    \item \textbf{Multi-Head Self-Attention Mechanisms}~\cite{vaswani2017attention}: Incorporates transformer-style attention:
    \begin{equation}
        \text{Attention}(Q, K, V) = \text{softmax}\left(\tfrac{QK^T}{\sqrt{d}}\right)V.
    \end{equation}
    These layers excel at capturing complex relationships between variables of different characteristics, potentially helping model the interactions between economic and climate dynamics.
\end{enumerate}

\subsection*{Architecture-Specific Enhancements}

We implement several specialized techniques to address particular challenges in our climate-economic model:

\begin{enumerate}
    \item \textbf{Boundary-Enforcing Constraints}: Custom output layers with sigmoid activations enforce known boundary constraints on policy functions, particularly for investment rates that must remain within physically meaningful bounds.
    \item \textbf{Adaptive Optimization Strategies}: We test multiple optimization algorithms (Adam~\cite{kingma2014adam}, AdamW~\cite{loshchilov2017decoupled}, SGD) with varying learning rates for value and policy networks.
    \item \textbf{Learning Rate Scheduling}: We test step scheduling and cosine annealing schedules with optional warm restarts to navigate the complex loss landscape.
    \item \textbf{Regularization and Stabilization}: We explore dropout~\cite{srivastava2014dropout}, weight decay, and gradient clipping to improve training stability.
\end{enumerate}

\subsection*{Hyperparameter and Training Details}\label{sec:hyperopt}
To ease the navigation of the complex design space of these adversarial framework, we design a comprehensive hyperparameter optimization, with the goal of finding combinations of hyperparameters that provide satisfactory solutions, thereby minimizing the distance with respect to the finite-differences solution $\mathcal{L}_{final}$. In terms of \textbf{Optimization}, we observe that AdamW with high weight decay (0.04), batch size of 4096, a gradient norm clipping of $1.0$, and betas of $\beta
_1 = 0.9, \beta_2 = 0.99$ provide satisfactory results. Learning rates should be different for each model ($0.001$ for $\mathcal{P}$, $0.0005$  for $\mathcal{V}$), and that learning rate schedulers do not provide consistent improvements. For each batch of samples, we train $\mathcal{P}$ for 20 steps, and $\mathcal{V}$ also for 20. Sampling is done uniformly across the space. In terms of \textbf{model sizes}, regardless of the layer design, we find that both models provide the best results when they have 5 hidden layers each, where the layers of $\mathcal{P}$ have 1024 neurons each, and $\mathcal{V}$ has 512. Dropout is set to $0$ as it failed to provide consistent gains. Note that this specific configuration works generally well for every combination of $\mathcal{V}$ and $\mathcal{P}$, but it may not be necessarily optimal.  It is, however, computationally unfeasible to test every possible combination of model design, size, optimization type, regularization, and every other hyperparameter choice. Our goal in this paper is to provide insights on this design space (with a particular focus on layer designs), rather than exploring it exhaustively. We train the models for 500 epochs, and run validation against the ground truth data at the end. In the following, we will provide experiments on how the architecture design impacts the final performance of our neural PDE solvers, using the training configuration we just described.

%% file: sections/results.tex
\begin{figure}[t!]
    \centering
    \includegraphics[width=1\columnwidth]{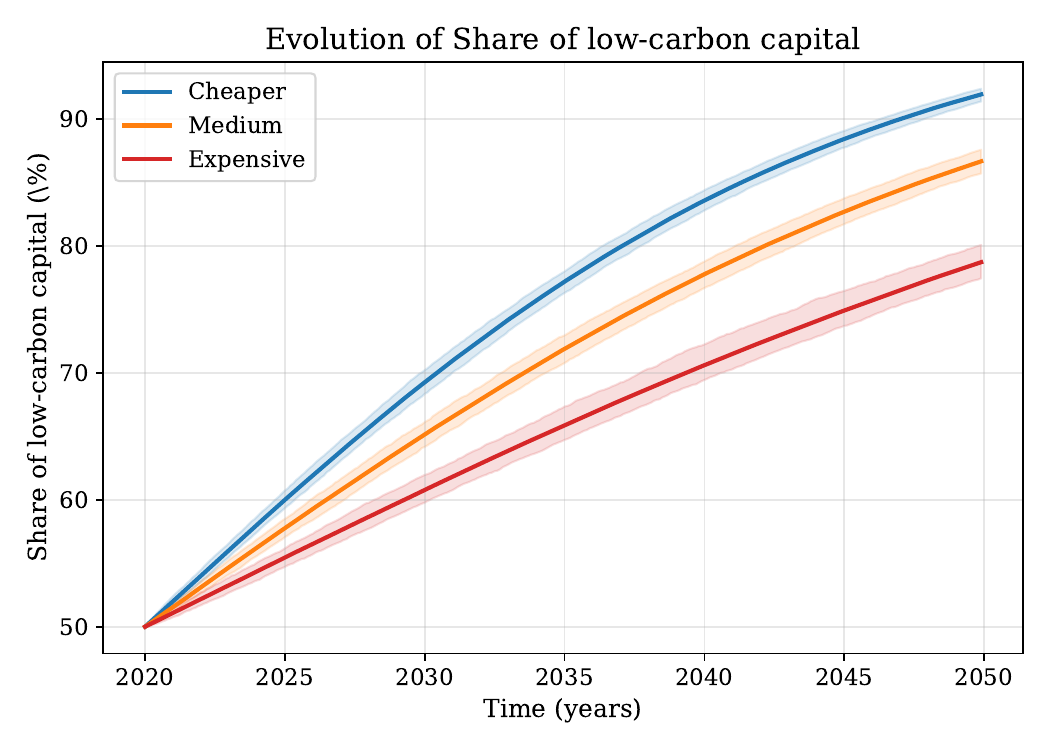}
    \caption{Comparison on the share of low-carbon capital (and 95\%) confidence ranges), emulated by our model on different prices assumptions.}
    \vspace{5mm}
    \label{fig:SL_comparison}
\end{figure}

\section{Results}\label{sec:results}

\paragraph*{Neural Network Comparisons} 
Our evaluation of neural architectures for climate-economic PDEs reveals several key insights into the relationship between network design, solution quality, and computational cost (Figure~\ref{fig:performance_comparison}).

The error analysis shows a clear performance stratification across architectures. MLP-based networks consistently achieve the lowest error rates when combined when MLP or recurrent models, outperforming specialized alternatives. This counter-intuitive finding suggests that the smoothness properties of climate-economic solution manifolds may be better captured by the universal approximation capabilities of MLPs than by the inductive biases of specialized architectures. Particularly striking is the poor performance of SIREN and Self-Attention networks, despite their theoretical advantages in representing high-frequency and complex components—indicating that such components may play a less significant role in our specific HJB formulation than anticipated. Interestingly, baseline DGM models can be improved with minor architectural adjustments, including the use of SiLU activations (DGM+S) and residual connections (DGM+SR), although these results are somewhat marginal. Models that operate on a Fourier space (FNO) generally provide good results as Value Networks, although inferior than recurrent layers like LSTM or GRU. Our best-performing method is trained on 16 minutes on a RTX 4090 GPU (compared to 2 hours for the FD). Using CodeCarbon~\cite{lacoste2019quantifying}, we measure an electricity consumption of 0.127646 kWh. 

Computational efficiency measurements reveal that architecture choice creates performance differences of up to 6× in training time, with MLPs demonstrating a clear advantage. The combined error-efficiency metric highlights the practical superiority of MLP-MLP configurations, achieving considerably better efficiency than attention-based alternatives. This suggests that the computational complexity of climate policy models might be effectively managed without resorting to more sophisticated architectures. 

Notably, our results challenge conventional practices regarding architecture selection for PDE solving. The significant performance gap between theoretical expectations and empirical results for specialized architectures (e.g., Self-Attention and SIREN) reveals a potential misalignment between these networks' inductive biases and the specific mathematical properties of climate-economic PDEs. The relatively smooth, low-frequency nature of optimal policy functions in our model appears to favor the simplicity and trainability of MLPs over architectures designed for more complex systems.
Our findings also demonstrate the asymmetric impact of value and policy network architecture choices. Value networks appear to have a greater influence on overall performance, with MLP value networks producing superior results regardless of policy network selection. This asymmetry provides practical guidance for computational resource allocation when designing neural PDE solvers for climate policy applications. These results show that careful design choices for the value function are more relevant than those for the policy model, which are comparably easier to train.

\paragraph*{Model Emulation} 
Once trained, we can use our model to test the impact of different hypothesis on the structure of the economy and its potential decarbonization.  On Figure~\ref{fig:SL_comparison}, we show an emulation (300 iterations) of our best performing model for the years 2020-2050, under different assumptions on the cost of low carbon capital. As shown, on more expensive scenarios, low carbon capital becomes a smaller share of the economy, thus hindering decarbonization efforts. These emulations also provide uncertainty estimates, and can enable more fine-grained policy designs, as they work on a fully continuous space. 

%% file: sections/conclusions.tex
\section{Conclusions}\label{sec:conclusion}
This paper has explored the impact of neural network architecture design on solving high-dimensional partial differential equations arising in climate-economic models. Our systematic evaluation of neural architectures demonstrates that design choices significantly affect both solution accuracy and computational efficiency when modeling climate policy under uncertainty. We examined various neural architectures including enhanced DGM variants, MLPs, RNN, SIREN, FNO, and attention-based architectures. Our analysis reveals that architecture selection plays an important role in capturing the unique characteristics of climate-economic PDEs. We also discovered that pairing different architectures for value and policy networks often yielded better results than using identical architectures for both components, which is the common practice. Our findings show that simpler neural networks offer better results both in accuracy and computational cost for this problem, paving the way for more scalable and performing neural PDE solvers in climate economics.

These methodological advances enable more sophisticated modeling of climate mitigation pathways under uncertainty, allowing policymakers to better understand the complex interplay between ambiguity aversion, technological transitions, and optimal policy design. Our approach helps bridge the gap between theoretical model complexity and practical policy analysis, by providing more accurate and computationally efficient solutions to high-dimensional climate-economic PDEs. Future work should extend our quantitative evaluation to additional model types and optimization algorithms, as well as explore higher-dimensional problems that incorporate additional mitigation technologies such as carbon capture. As climate policy models continue to grow in complexity to reflect real-world challenges, the architectural insights provided in this paper will become increasingly valuable for computational climate economics.